\pdfoutput=1

\documentclass[11pt]{article}

\usepackage{ACL2023} 

\usepackage{times}
\usepackage{latexsym}

\usepackage[T1]{fontenc}

\usepackage[utf8]{inputenc}

\usepackage{microtype}

\usepackage{inconsolata}

\usepackage{graphicx}

\usepackage{hyperref}

\usepackage{multirow}
\usepackage{tabularx}
\usepackage{float}

\title{Towards Automatic Construction of Filipino WordNet: Word Sense Induction and Synset Induction Using Sentence Embeddings}

\author {
\textbf{Dan John Velasco\thanks{\;\,Corresponding author. Direct all communications to: \texttt{dan\_velasco@dlsu.edu.ph}}}$^{*1}$\quad
\textbf{Axel Alba}$^{1}$\quad
\textbf{Trisha Gail Pelagio}$^{1}$\quad
\textbf{Bryce Anthony Ramirez}$^{1}$\quad
\\
\textbf{Unisse Chua}$^{1}$\quad
\textbf{Briane Paul Samson}$^{1}$\quad
\\
\textbf{Jan Christian Blaise Cruz\thanks{\;\,Work done while at De La Salle University, Manila. }}$^{\dag 2}$\quad
\textbf{Charibeth Cheng}$^{1}$\quad
\
\\
$^1$De La Salle University, Manila \;
$^2$Samsung R\&D Institute Philippines \;
}

\begin{document}
\maketitle
\begin{abstract}
Wordnets are indispensable tools for various natural language processing applications. Unfortunately, wordnets get outdated, and producing or updating wordnets can be slow and costly in terms of time and resources. This problem intensifies for low-resource languages. This study proposes a method for word sense induction and synset induction using only two linguistic resources, namely, an unlabeled corpus and a sentence embeddings-based language model. The resulting sense inventory and synonym sets can be used in automatically creating a wordnet. We applied this method on a corpus of Filipino text. The sense inventory and synsets were evaluated by matching them with the sense inventory of the machine translated Princeton WordNet, as well as comparing the synsets to the Filipino WordNet. This study empirically shows that the 30\% of the induced word senses are valid and 40\% of the induced synsets are valid in which 20\% are novel synsets.
\end{abstract}

\section{Introduction}
In Natural Language Processing (NLP), a language resource such as the Princeton WordNet or PWN \cite{Miller1993} has been widely used in several NLP tasks, including sentiment analysis \cite{kumar-etal-2018-knowledge}, word sense disambiguation \cite{mizuki-okazaki-2023-semantic}, and machine translation \cite{ngo-etal-2019-overcoming}. A wordnet contains \emph{word senses} which is a discrete representation of one aspect of the meaning of a word \cite{slp3}, \emph{synsets} or synonym sets which are sets of near-synonyms, \emph{word definitions},  \emph{example sentences},  \emph{part-of-speech}, and  \emph{semantic links} (e.g synonyms, hyponyms, and meronyms). Such lexical resource can be useful for language documentation, language preservation, and linguistics research.

For the Filipino language, there is the Filipino WordNet or FilWordNet \cite{Borra2010} which was manually constructed by translating PWN senses to Filipino and adding unique Filipino words that are not in the PWN. Overall, it consists of 13,539 unique words and 9,519 synsets. Each synset in the FilWordNet includes semantic information such as the word's definition, part-of-speech, word senses, and Suggested Upper Merged Ontology terms \cite{Niles2001}.

However, the FilWordNet was not updated since its creation. It does not contain new words and senses that emerged in the Filipino language, most notably the colloquial words used in digital media and platforms. For comparison, the University of the Philippines Diksiyonaryong Filipino, a monolingual Filipino dictionary, contains over 200,000 word senses \cite{ManilaBulletin}, while FilWordNet only documents 15,929 word senses. One probable reason for FilWordNet not being updated is the expensive cost of manually creating and updating a wordnet in terms of time and resources.

Advancements in NLP can be leveraged to make wordnet construction more efficient. In this study, we propose a method for automatic generation of sense inventory and synset inventory, remedying the inefficiencies in the pure manual process. Our approach only requires an unlabeled corpus and a sentence embeddings-based language model. Human supervision is not required during the generation process. Our technique is language-agnostic, but for this work, we use Filipino as a case study to produce sense inventories and synset inventories, which are among the core components of a wordnet.

\section{Related Work}
\subsection{Automatic Wordnet Construction}

According to \citet{Vossen1998}, wordnets are typically built using two approaches: (1) the merge approach, where lexicographers manually construct the wordnet, and (2) the expansion approach, which utilizes an existing reference wordnet as a guide in constructing the new wordnet. There is a trade-off between quality and cost with merge approach producing higher quality wordnets at a high cost and with expansion approach producing relatively lower quality but at a significantly lower cost \cite{bhattacharyya-2010-indowordnet}. Sense inventory and synsets are one of the core components of a wordnet and the focus of this paper. Different approaches to induce these components will be discussed in Sections \ref{sec:rrl_synset_induction} and \ref{sec:rrl_sense_induction}.

\subsubsection{Synset Induction}
\label{sec:rrl_synset_induction}

According to the recent survey paper on automatically constructed wordnets by \citet{neale-2018-survey}, most works follow the expansion approach which uses a reference wordnet (usually PWN) and a lexical resource such as encyclopedia \cite{casado2005}, bilingual dictionary or machine translator \cite{barbu-mititelu-2005-case, montazery-faili-2010-automatic, khodak-etal-2017-automated, rahit-etal-2018-banglanet}, and parallel corpus \cite{oliver-climent-2014-automatic} to construct synsets for their target language and often used in conjunction with word embeddings. These approaches are efficient but have a hard requirement of having a reference wordnet and a lexical resource. Our approach differs from these works by not requiring a reference wordnet or a lexical resource to induce synsets.

\subsubsection{Word Sense Induction}
\label{sec:rrl_sense_induction}

Most works on Word Sense Induction (WSI) are not directly motivated in the automatic construction of wordnet but are still related. WSI is an essential component of automatic wordnet construction without a reference wordnet and a lexical resource, which is the goal of this paper. Recent works on WSI use clustering of sense embeddings derived from word embeddings \cite{song-etal-2016-sense}, lexical substitutions derived from language models \cite{amrami2019, arefyev-etal-2019-combining}, word co-occurence graphs \cite{bekavac-snajder-2016-graph}, sparse coding and sentence embeddings \cite{khodak-etal-2017-automated}, and clustering of sentence embeddings \cite{tallo2020}. Our work is closely related to \citet{tallo2020} where we learn sentence embeddings following SBERT \cite{reimers-gurevych-2019-sentence} but for Filipino and then induce senses by clustering sentence embeddings. Another difference is that our work used Affinity Propagation\footnote{\url{https://scikit-learn.org/stable/modules/generated/sklearn.cluster.AffinityPropagation.html}} which does not require knowing the number of clusters beforehand while \citet{tallo2020} used K-Means Clustering which requires specifying the number of clusters in advance.

While many automatic wordnet construction approaches rely on a reference wordnet or other lexical resources, our method eliminates the need for these references. This is particularly beneficial for low-resource languages, where such lexical resources are often lacking. Our approach offers greater accessibility to low-resource languages, enabling the induction of culturally and regionally specific concepts unique to each language. Moreover, it avoids the distraction and constraints imposed by reference wordnets in other languages, as it does not depend on words with direct translations. For instance, consider the word 'gigil,' which describes the trembling response to situations overwhelming our self-control, a concept not directly translatable to English. It's likely that this word would not be covered in the English wordnet, as it is unique to the Filipino language.

\subsection{Sentence Embeddings}
Sentence embeddings maps a sentence to a vector space such that semantically similar sentences are close in proximity. A simple method for creating sentence embeddings is to average token embeddings of a sentence \cite{Arora2017}. However, this approach ignores word interactions within a sentence. This was addressed by \citet{reimers-gurevych-2019-sentence} where they used BERT in a siamese and triplet network setup to generate meaningful sentence embeddings by training on SNLI \cite{bowman-etal-2015-large} and MultiNLI \cite{williams-etal-2018-broad} datasets which achieved state-of-the-art results in Semantic Textual Similarity (STS) tasks at time of publication. In recent years, unsupervised methods for learning sentence embeddings have been proposed such as TSDAE \cite{DBLP:journals/corr/abs-2104-06979}, SimCSE \cite{gao-etal-2021-simcse}. and DiffCSE \cite{chuang-etal-2022-diffcse} but for this paper, we only experimented on the approach following \citet{reimers-gurevych-2019-sentence}. We apply sentence embeddings for WSI through clustering. Our method assumes that semantically similar sentences exhibit proximity to one another, rendering them suitable for clustering and WSI.

\section{Datasets}
\subsection{Corpus}
Filipino texts were collected from various domains such as news sites, books, social media, online forums, and Wikipedia to build a diverse corpus that represents wide range of writing styles using the Filipino language. Table \ref{tab:corpus_stats} shows the number of words and the average sentence length for each domain. Personal identifiers, symbols, and entities that are irrelevant to the study, such as emails, emojis, links, hashtags, non-alphanumeric symbols were removed. The documents were splitted into sentences using NLTK sentence tokenizer\footnote{\url{https://www.nltk.org/api/nltk.tokenize.html}}. The following tokens are treated as sentence delimiters: question marks (?), exclamation points (!), ellipses (...), periods (.), and line breaks. The language of each sentence were identified using fastText \cite{joulin2016bag}. For the purpose of this work, we only retained sentences that were identified as Tagalog/Filipino. The resulting corpus contains around 997 million words of which 5.37 million words are unique. All of this unique
words should not be assumed to be valid words since it may contain non-natural language such as symbols.

For sources that are small enough and do not update frequently, such as Google Books and Bible\footnote{\url{https://www.wordproject.org/bibles/tl/index.htm}}, they were downloaded manually. For YouTube, the official YouTube Data API\footnote{\url{https://developers.google.com/youtube/v3}} was used. We curated a set of YouTube channels of Filipino content creators and collected all the comments of their videos. The curated list of YouTube channels can be found at Appendix \ref{sec:youtube_channels}. For Twitter, the official Twitter API with Academic Research access\footnote{\url{https://developer.twitter.com/en/use-cases/do-research/academic-research}} was used. We only collected tweets with 'fil' language tag which is the BCP47 language tag for Filipino. For Reddit, the Pushshift API\footnote{\url{https://reddit-api.readthedocs.io/en/latest/}} was used to collect posts and responses in subforums or what they call subreddits. We curated a list of subreddits where Filipinos interact. The curated list of subreddits can be found at Appendix \ref{sec:subreddits}. For Tagalog Wikipedia, a dump file from Wikimedia\footnote{\url{https://dumps.wikimedia.org/tlwiki/latest/}} were downloaded and preprocessed. For news sites, web scrapers were built for each website using Scrapy\footnote{\url{https://scrapy.org/}} for sites with pagination and Selenium\footnote{\url{https://www.selenium.dev/}} for sites with infinite scrolling. For a complete list of sources, refer to Appendix \ref{sec:data_sources}.

\begin{table}[]
\begin{tabularx}{\columnwidth}{|l|X|l|}
\hline
Domain & Total Words & \begin{tabular}[c]{@{}l@{}}Mean \\ Sentence \\ Length\end{tabular} \\ \hline
Books & 8,923,001 & 23 \\ \hline
News & 94,466,149 & 20 \\ \hline
Online Forums & 92,342,868 & 15 \\ \hline
Social Media & 792,778,696 & 12 \\ \hline
Wikipedia & 9,177,188 & 32 \\ \hline
\end{tabularx}
\caption{Corpus statistics showing total words and mean sentence length of each domain.}
\label{tab:corpus_stats}
\end{table}

\subsection{Filipino WordNet (FilWordNet)}
\label{FilWordNet}
The Filipino WordNet \cite{Borra2010} was manually constructed by translating the Princeton WordNet \cite{Miller1993} and then adding unique Filipino words. FilWordNet contains 13,539 unique words and 9,519 synsets. Our approach does not require a reference wordnet and FilWordNet was only used for evaluation purposes. All experiments used the vocabulary of FilWordNet as seed words, with a few exceptions. FilWordNet words that have less than 20 example sentences in our corpus were excluded. FilWordNet words that start with an uppercase letter were excluded to remove proper nouns. FilWordNet words with only 2 letters or less were also excluded. Lastly, FilWordNet words that represent a number were also removed. From the initial 13,539 words of the FilWordNet, only 2,598 remained after filtering and were used as seed words. Unless otherwise stated, all experiments used the filtered seed words from FilWordNet. We restrict this procedure to direct comparisons.  In practice, the seed words can originate from the target corpus's vocabulary or its set of distinct words. Nonetheless, based on the specific use case, additional preprocessing may be required, as the corpus vocabulary may include non-natural language tokens, such as emojis or symbols.

\begin{figure*}[!htbp]
        \centering
        \includegraphics[width=\textwidth]{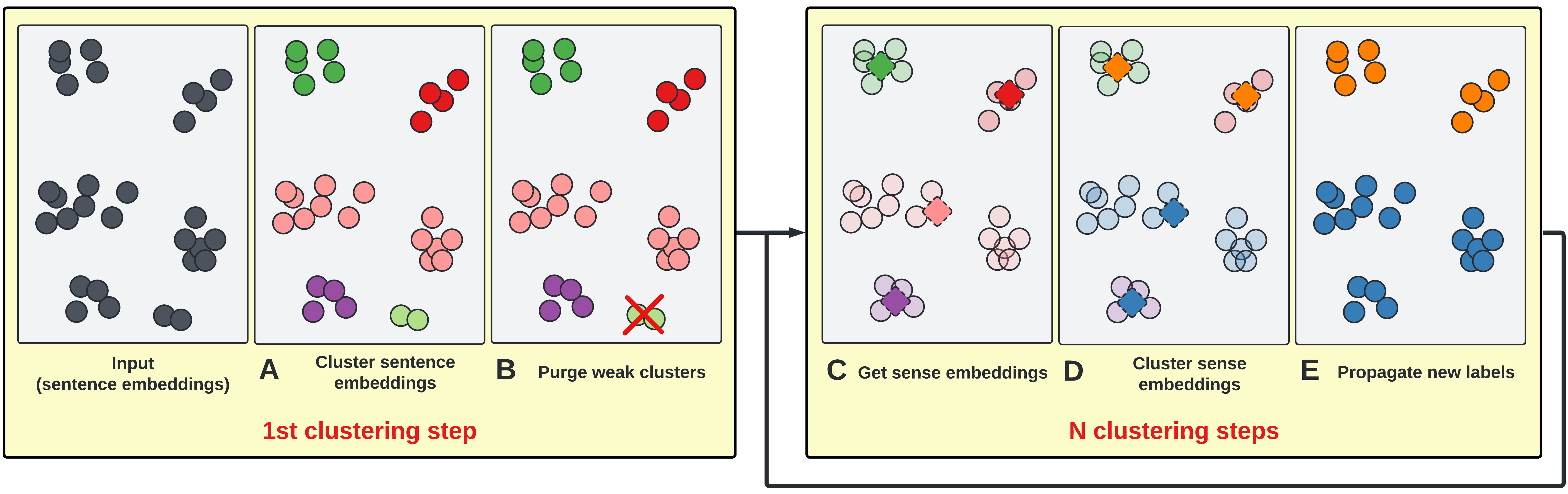}
        \caption{\textbf{N-STEP Clustering}. First clustering step (\textbf{A} to \textbf{B}) clusters sentence embeddings (\textbf{A}) and purges weak clusters (\textbf{B}). N clustering steps (\textbf{C} to \textbf{E}) is iterative. (\textbf{C}) Computes sense embeddings followed by (\textbf{D}) clustering, effectively reducing the points to cluster. (\textbf{E}) Propagates the new labels to its cluster members and goes back to (\textbf{C}).}
        \label{fig:nstep}
\end{figure*}

\section{Methodology}
\subsection{Language Model Training}
The training goes through two phases: (1) Masked Language Modeling (MLM) and (2) Contrastive Learning. In the first phase, we adapt the language model to our corpus by finetuning the pretrained RoBERTa for Filipino \cite{cruz-cheng-2022-improving} on our corpus for 1,176,690 steps (equivalent to 10 epochs) with a maximum learning rate of 5e-5 and then linearly decayed. Doing so can improve representations especially for social media writing styles since our corpus mostly contains social media data and the pretrained RoBERTa was only trained on more formal writing styles such as news data and Wikipedia articles. The AdamW optimizer \cite{loshchilov2019decoupled} was used with the following hyperparameters: $\beta_1$ = 0.9, $\beta_2$ = 0.999, and $\epsilon$ = 1e-6. The pretrained RoBERTa was finetuned on Google Compute servers with TPUv3-8 accelerators.

In the second phase, we finetune the model from the first phase on NewsPH-NLI dataset \cite{newsphnli2021} in a contrastive learning setup following the work of \citet{reimers-gurevych-2019-sentence} but used Multiple Negatives Ranking Loss\footnote{\url{https://www.sbert.net/docs/package_reference/losses.html}} instead of the Softmax Loss used in the original work. We used mean of contextual token embeddings as pooling method. The positive entailment pairs used in training are 237,679 sentence pairs. The model was optimized with AdamW optimizer using the following hyperparameters: \emph{epochs} = 1, \emph{max learning rate} = 2e-5, \emph{max sequence length} = 128, and \emph{batch size} = 16. The learning rate gradually increases up to max learning rate for the first 10\% of training and then linearly decayed. We will refer to this model as \emph{filipino-sentence-RoBERTa}\footnote{\url{https://huggingface.co/danjohnvelasco/filipino-sentence-roberta-v1}}. The model was trained on one NVIDIA GeForce RTX 3060Ti GPU.

\subsection{Word Sense Induction using N-STEP Clustering \label{WSI_section}} 

This method is iterative. For each word in the list of seed words, the Sentence Sampling (Section \ref{sentence_sampling}) and N-STEP Clustering (Section \ref{nstep_clustering}) are repeated.

\subsubsection{Sentence Sampling}
\label{sentence_sampling}
The first step in WSI is to sample sentences from a corpus. For each word, sentences are sampled from each domain in the corpus (e.g. books, news sites, etc.) for sentences containing the target word through substring matching wrapped with whitespaces (e.g. " hot "). The sample sentences are splitted by whitespace to apply windowing with the target word at the center\footnote{refer to ~\nameref{limitation:embeddings} section on why windowing was done}. For this study, a fixed window size of 4 for both left and right was set through trial and error and was not extensively studied. Due to time and memory constraints, a maximum sample size of 1,000 sentences per source was set. The resulting sample sentences for a target word will be encoded to sentence embeddings using filipino-sentence-RoBERTa.

\subsubsection{N-STEP Clustering}
\label{nstep_clustering}
Applying clustering once on a large sample of sentences results to hundreds of clusters. Through manual inspection, we observed that many clusters are redundant i.e. multiple clusters pertain to the same sense. To address this, we proposed \textbf{N-STEP Clustering} (see Figure \ref{fig:nstep}) approach for WSI which performs clustering multiple times to reduce redundant clusters. The clustering algorithm used is \emph{Affinity Propagation} (AP) which does not require knowing the number of clusters in advance. We propose two (2) ways to implement N-STEP Clustering.

\textbf{Fixed N-STEP Clustering.} The number of clustering steps \emph{N} is fixed and must be set in advance. It works by clustering the sentence embeddings only at the first clustering step. In the succeeding steps until step \emph{N}, these clusters are iteratively merged together by clustering the \emph{sense embeddings} or the centroids (see Figure \ref{fig:nstep}). The use of centroids to represent a sense is similar to \citet{hu-etal-2019-diachronic, amrami2019}. 

\textbf{Dynamic N-STEP Clustering.} The process is similar to Fixed N-STEP Clustering. The only difference is that the number of clustering steps are dynamically set for each target word by maximizing the silhouette score. Silhouette score \cite{ROUSSEEUW198753} is a metric for evaluating the quality of clusters by looking at the closeness of each members within a cluster and the separation among the clusters. The silhouette score ranges from -1 to 1, where higher values indicates higher quality clusters. This allows for different number of clustering steps for each word which highly depends on the distribution of the sample sentences. N-STEP Clustering stops at the highest silhouette score after not improving for two (2) consecutive clustering steps. In the first clustering step, weak clusters were removed. Weak clusters are defined as clusters with less than \emph{k} members as they are often just noise. The idea of removing weak clusters is similar to \citet{amrami2019} but instead of merging the weak clusters with strong cluster, we simply remove the weak cluster. For our dataset, we set \emph{k}=3 but it can be adjusted depending on the dataset.

In this work, we used Fixed 3-STEP Clustering because it outperformed Dynamic N-STEP Clustering (see Appendix \ref{sec:3step_experiment_section}) on our dataset. This may not
generalize to other datasets and we suggest using
Dynamic N-STEP Clustering as a starting point. For all clustering steps, the damping parameter of the AP algorithm is set to 0.5. The justifications behind this value are explained in Appendix \ref{sec:damping_experiment_section}. Currently, there are no automatic methods to tune these parameters, but Appendix \ref{sec:3step_experiment_section} to \ref{sec:distance_thres_experiment_section} can provide some ideas to help future researchers find the best parameters for their own application. 

\subsection{Synset Induction}
In synset induction, all senses in the sense inventory will be clustered by clustering the sense embeddings. This is in contrast with WSI where clustering happens for each word. For synset induction, it is ideal to cluster only the similar senses without forcing other senses to belong to a synset. To achieve this effect, we used Agglomerative Clustering\footnote{\url{https://scikit-learn.org/stable/modules/generated/sklearn.cluster.AgglomerativeClustering.html}}, where the cosine distance threshold can be set. In this paper, the distance threshold has been set to 0.12, meaning only senses with a cosine distance of 0.12 or less will be clustered with other senses. The justification behind this threshold value are explained in Appendix \ref{sec:distance_thres_experiment_section}.

\section{Evaluation}
\label{experiments_sections}

\subsection{Sense Inventory Evaluation}

\begin{figure}[!ht]
\begin{center}
\includegraphics[width=\linewidth]{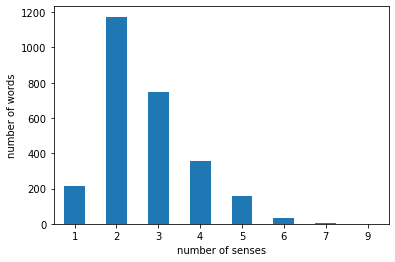} 
\caption{The distribution of the number of senses per word shows that most induced words have 2 senses. Words with higher sense count become less frequent as the number of senses per word increases. This also shows that it can induce single senses.}
\label{fig:sense_count_distribution}
\end{center}
\end{figure}

The induced sense inventory contains 2,598 unique words and 7,177 senses. Most words have 2-3 senses and some words can have up to 9 senses (see Figure \ref{fig:sense_count_distribution}). We evaluate the sense inventory in a Word Sense Disambiguation (WSD) setup adapted from \citet{hu-etal-2019-diachronic}, where sentences from a reference sense inventory are tagged using our sense inventory. The sense embedding with the highest cosine similarity with the input sentence and is greater than or equal to the threshold is chosen as the sense for that input sentence. The induced senses were classified as valid if it is used as a tag in WSD at least once. In this paper, the average cosine similarity (0.63) was used as threshold for simplicity and was not extensively studied.

Evaluating in this setup requires a reference sense inventory but since the FilWordNet has no example sentences and there is no other existing lexical resource in Filipino that can be used to evaluate the sense inventory, we resorted to translating sentences from PWN as our evaluation data. For this step, we used the Google Translate API as a translation tool\footnote{\url{https://cloud.google.com/translate}} and NLTK\footnote{\url{https://www.nltk.org/}} to access PWN. The Filipino words were translated to English and were used to retrieve PWN senses. The example sentences of the PWN were translated back to Filipino. However, not all of the translations of the induced senses can be found in the PWN because some Filipino words do not have a direct English translation. In addition, some of the PWN entries were discarded because the English to Filipino backtranslation does not contain the target word in Filipino. As a result, the final evaluation data can only evaluate 23\% of the unique words (601 out of 2,598 words) of the generated sense inventory. 

The results show that around 30\% (552 out of 1,864) of the induced senses were matched with at least one of the senses from the PWN. Upon manual inspection of the remaining 70\%, it is observed that most of the induced senses are either: (1) a sense that are not found in the PWN or English language, (2) clusters with multiple senses inside, (3) redundant clusters that represent an already existing sense. Thus, not finding the induced sense in the PWN does not necessarily mean that the induced sense is invalid. Further analysis, such as a human evaluation, is needed to determine whether the unmatched senses are incorrect. We leave this for future work.

\subsection{Synset Evaluation}
\subsubsection{Automatic Synset Evaluation}
We evaluate the correctness of the induced synsets by comparing it with FilWordNet synsets using Jaccard Index where a score of 1 means exact match and 0 means exact mismatch. For fair comparison, FilWordNet synsets with only one sense were excluded. While our sense inventory and FilWordNet share the same vocabulary, some words are not found in the corpus used for WSI. Hence, FilWordNet synsets containing words that are not in the vocabulary of our sense inventory were excluded. After filtering, the total number of FilWordNet synsets is down from 9,519 to 639 synsets. This evaluation is comparing 1,109 induced synsets to 639 FilWordNet synsets.

\begin{figure}[!ht]
\begin{center}
\includegraphics[width=\linewidth]{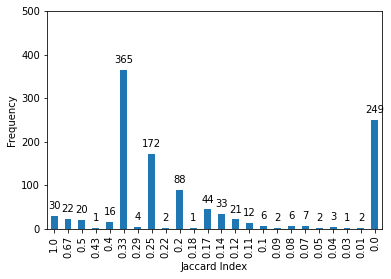} 
\caption{Jaccard Index of the induced synsets and FilWordNet synsets.}
\label{fig:jaccard_scores}
\end{center}
\end{figure}

Figure \ref{fig:jaccard_scores} shows that 3\% (30 synsets) of the induced synsets are exact match in FilWordNet. The Jaccard Index with the highest frequency is 0.33 which comprised of 33\% (365 synsets) of the induced synsets. On the other hand, 22\% (249 synsets) are exact mismatch. For this experiment, a high Jaccard Index is preferred but getting low scores does not necessarily mean that the induced synsets are of poor quality. There could be several reasons that could lead to a low Jaccard index score such as the discovery of new senses and synsets that are not in FilWordNet. With these results, a human evaluation is needed to better validate the quality of the induced synsets that is independent from FilWordNet synsets.

\subsubsection{Manual Synset Evaluation} 
To give a better assessment of synset quality that is independent from FilWordNet synsets, a human evaluation is needed. This was done by scoring the synsets with the percentage of correct senses in the synset. For example, given \textit{A = \{'big', 'large'\}}, it would be labeled with a score of 1.0. On the other hand, given \textit{B = \{'hard', 'difficult', 'challenging', 'large', 'blue'\}}, it would be labeled with a score of 0.60 as three out of the five words (\textit{'hard', 'difficult', 'challenging'}) are synonyms while the other words are irrelevant. To simplify the annotation, synsets with more than 10 senses were labelled as 0 since most FilWordNet synsets only contains 1 to 4 words inside the set. In addition, synsets that contain only 2 senses can only be labeled as 1 or 0. 

\begin{figure}[!htbp]
    \centering
    \includegraphics[width=\linewidth]{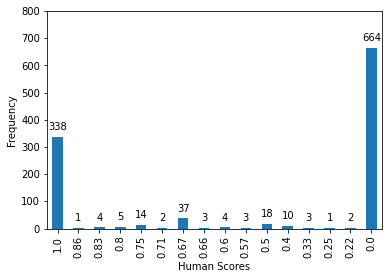}
    \caption{Frequency distribution of human scores from manual synset evaluation.}
    \label{fig:manual_scores6}
\end{figure}

Figure \ref{fig:manual_scores6} shows the distribution of scores made by humans. The extreme values of 1.0 and 0.0 obtained the highest number of synsets. Specifically, 30\% (338 synsets) of the induced synsets are 100\% correct and around 10\% (107 synsets) are partially correct. This is a big difference from the automatic results which shows that only 3\% of the induced synsets are 100\% correct. On the other hand, synsets with a score of 0.0 remained to have the highest number of counts which is 60\% (664 synsets) of the induced synsets.

\subsubsection{Types of Correct Synsets}
\begin{figure}[!htbp]
        \centering
        \includegraphics[width=\linewidth]{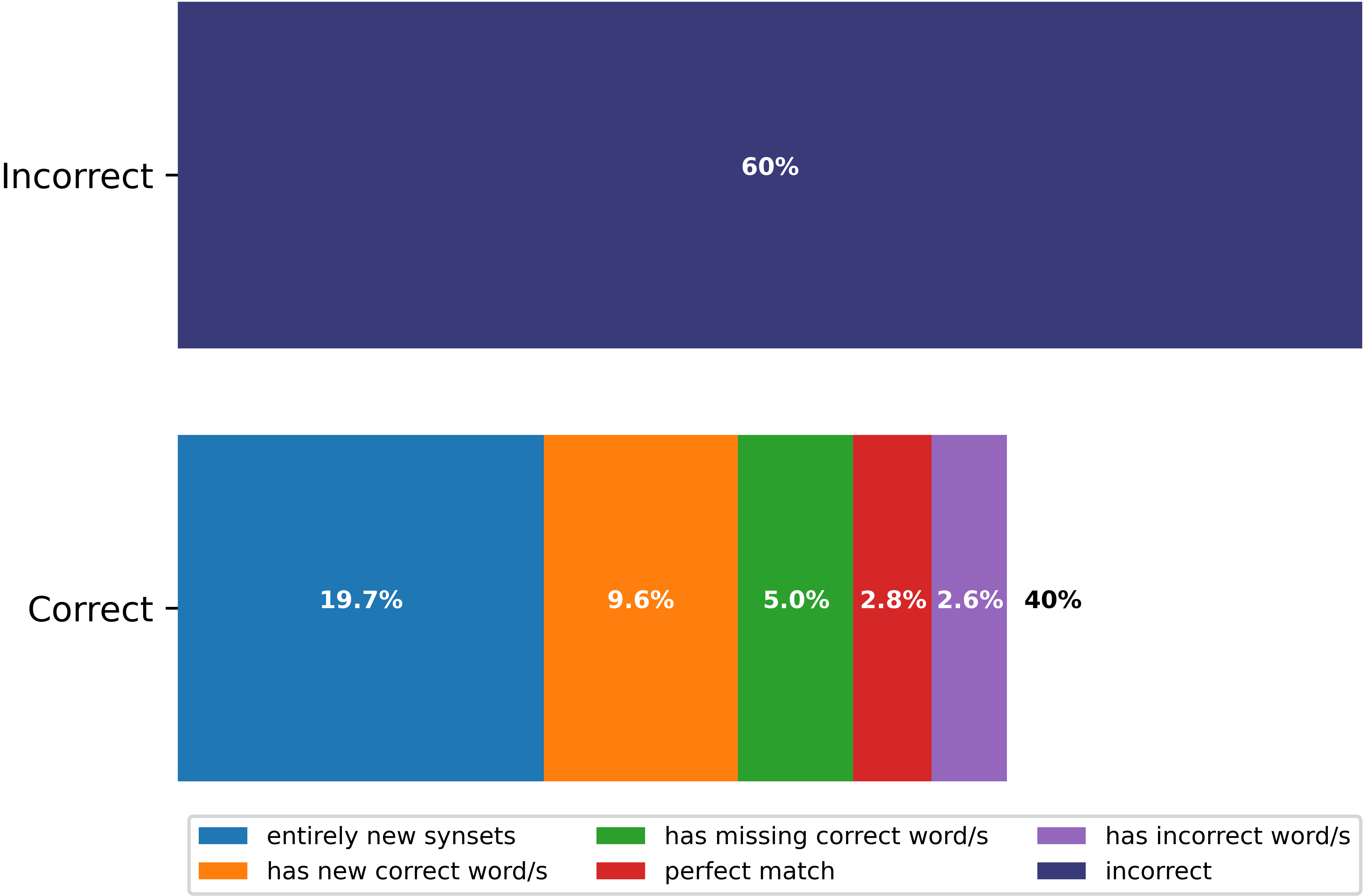}
        \caption{Types of correct synsets. The top two most frequent types shows the potential of our approach in synset discovery.}
        \label{fig:correct_types}
\end{figure}

To give a better view of what the correct and partially correct synsets are, we manually identified the types of correct synsets (see Figure \ref{fig:correct_types}). The top two most frequent type of correct synset is "entirely new synsets" (19.7\%) and "has new correct word/s" (9.6\%), followed by "has missing correct word/s" (5.0\%), "perfect match" (2.8\%), and "has incorrect word/s" (2.6\%). The top two most frequent type relates to inducing new synsets which indicates the potential of our approach in the discovery of synsets in an unsupervised manner.

Overall, the synset evaluation demonstrates the potential of our approach to synset induction which could greatly remedy the inefficiencies in the pure manual process of constructing a wordnet.

\section{Conclusions and Future Work}
In this work, we proposed (1) N-STEP clustering of sentence embeddings for word sense induction and (2) clustering of sense embeddings or centroids for synset induction. Both can be used towards automatic construction of a wordnet. The advantage of our method is that the generation is fully unsupervised and has only two requirements: unlabeled corpus and sentence embeddings-based language model, which are both accessible even for low-resource languages.

For word sense induction, the results showed that approximately 30\% (552 out of 1,864) of the induced senses are valid. Upon manual inspection of the remaining 70\%, it is observed that most of the induced senses are either: (1) sense that is not found in the PWN or English language; (2) clusters with multiple senses inside; (3) redundant clusters that represent an already existing sense. For synset induction, the results showed the potential of our approach to synset induction given that 40\% of the induced synsets are correct (30\% are perfectly correct and 10\% are partially correct). Among the correct synset types, the top two most frequent type pertains to discovering new synsets which demonstrates the potential of our approach in sense and synset discovery in an unsupervised manner. 

For future work, an evaluation set for WSI in Filipino similar to SemEval 2010 Task 14 \cite{manandhar-etal-2010-semeval} can be developed to get a better assessment of WSI performance. Future work may also look into the automatic induction of semantic links (e.g synonyms, hyponyms, and meronyms) without a reference wordnet to enable efficient construction of wordnets in low-resource languages. Further improvements to our work may come from the knowledge representation (e.g. better sentence embeddings or directly learning sense embeddings for WSI and synset induction). The use of unsupervised approaches to learning sentence embeddings \cite{gao-etal-2021-simcse, chuang-etal-2022-diffcse} can also be explored.

\section*{Limitations}
\subsection*{Evaluating on machine translated Princeton WordNet}
The evaluation on word sense induction was done on machine translated Princeton WordNet due to the lack of language resource in Filipino suitable for evaluation. There are several issues with this setup: (a) machine translation can introduce more errors to the evaluation; (b) some example sentences from Princeton WordNet are too short to disambiguate the word sense which influences the evaluation through word sense disambiguation; (c) some senses in the Filipino language may not appear in the English language (vice versa) which can lead to high false negatives.

\subsection*{Undesirable property of sentence embeddings for synset and word sense induction}
\label{limitation:embeddings}
While sentence embeddings shows its potential application for automatic wordnet construction, our approach produced a large number of incorrect synsets (60\%). The clusterability of the representations largely influences the performance on word sense induction and synset induction. One instance that highlights this problem is the problem of collocating words. When using sentence embeddings for WSI, words that do not necessarily have the same meaning would be closely related in terms of context. For example, the collocated words \emph{angat (raise)} and \emph{bandera (flag)} have different meanings but they were clustered as a synset since they are close in the embeddings space because they often appear together in the same context (e.g. "to raise the flag").

Using sentence embeddings in clustering is not always a robust way to identify senses that are used in very diverse contexts. For example, the word \emph{bank} as in the financial institution, can appear in very different contexts. The assumption of sentence embeddings is that sentences with similar meaning are close together in the vector space. Which means, even if the sample sentences only have one usage of the word \emph{bank} but it appears in very diverse contexts, there is the tendency to produce multiple clusters even though they represent the same meaning. This happens because the representations represents the semantics of the sentence itself rather than the sense usage. This is the intuition on why we implemented windowing before word sense induction. This also explains the high number of redundant clusters in word sense induction.

\section*{Acknowledgements}
This research was funded by the Philippine Department of Science and Technology - Philippine Council for Industry, Energy and Emerging Technology Research and Development. The authors would also like to acknowledge TensorFlow Research Cloud (TRC) program which made training models on TPUs possible.

\bibliography{anthology,custom}
\bibliographystyle{acl_natbib}

\appendix
\section{YouTube Channels}
\label{sec:youtube_channels}

The list of YouTube channels were manually curated. The criteria for YouTube channel selection is that the channel owners are Filipino, preferably, speaking in Filipino in their videos to increase the likelihood that the comments are in Filipino or mostly written by Filipinos. The following is the complete list of YouTube channels we collected data from: cong tv, slater young, kryzzzie, alodia gosiengfiao, rei germar, ry velasco, anna cay, anne clutz, anneclutzvlogs, itsclaudineco, abs cbn news, abs cbn entertainment, alliana dolina, donnalyn bartolome, akosi dogie, carlo ople, oliver austria, kenny manalad, doc willie ong, rendon labador, kiko matos, mika salamanca, presello, alex gonzaga official, h2wo, shinejonzoief2p, junnie boy, benedict cua, toni gonzaga studio, tryke gutierrez, reccreate, bermor, rj jacinto, xtian c, fliptopbattles, panlasang pinoy, bea alonzo, karl jolice tv, dubstep, front seat foodies, nisky, ninong ry, gma public affairs, raffy tulfo in action, bonoy pinty gonzaga, leti sha, wish 1075, homesearch philippines, kiyo, ben\&ben, vice ganda, jah de dios, araw na itim tagalog animated horror, paolul, mavs phenomenal basketball, jonahrenz jacob, sb19 official, karla cayabyab, mocha uson official, leni robredo, jinkee pacquiao, team pacquiao, bongbong marcos, karen davila, rappler, isko moreno domagoso, kuya kim atienza, euphoniaco tv, kathryn bernardo, viva records, dr. vicki belo, schizo pelma tv, joyce pring, nicole alba, juancho trivino, yow, latest scoop, mimiyuuuh, marjorie barretto, tiktok pinas, tiktok philippines, erwin tulfo, luis azcona sharlene menu, pinoy knows tech, eat bulaga, kusina ni lola, beebuyog, alvin tries tech, unbox ph, sulit tech reviews, choox tv, pepesan tv, olip tv, khent bernardo, the juans, clr, richard gomez, pinoy mystery channel, tinagalog, historya channel, hazel quing, ivana alawa, niana guerrero, zeinab harake, jelai andres, ramon bautista, empoy official, kuya jobert tv, gretchen ho, tv5 philippines, luis manzano, ogie diaz, love marie escudero, master long mejia, viy cortez, mentot, kevin hermosada, datwo, boss keng, vien, pat velasquez, beks battalion, youlol, jamill, chad kinis, riva quenery, wilbert tolentino, rana harake, daisy lopez, tita krissy achino, jamichtv, lloyd cafe cadena, limuel huet, mary bautista, michelle dy, kathsepaganvlogs, mikhaela cruz, peachytwice, john ryan santos, i can see your voice ph, the boy abunda talk channel, pinoy big brother, unbox diaries, the voice of the philippines.

\section{Subreddits}
\label{sec:subreddits}
The list of subreddits were manually curated. The criteria for subreddit selection is that most of the posters or members under that subreddit should be Filipinos. We determine this by looking for subreddits with the "PH" prefix or suffix in the subreddit name. We also included subreddits whose name represents a concept that only makes sense in the Philippine context (e.g. dlsu, ADMU, Tomasino, peyups subreddits refer to the Philippine universities). The following is a complete list of subreddits that we collected data from: phinvest, Philippines, NintendoPH, peyups, ADMU, dlsu, Tomasino, filipinofood, OldSchoolPH, PampamilyangPaoLUL, 3DSPH, beautytalkph, Bicol , BPOinPH, cagayandeoro, Cebu, davao, FilipinoFreethinkers, FilmClubPH, Iloilo, ilustrado, indiemusicph, KakaiBalita, Kwaderno, LoLPHSubreddit, mnl, opm, palawan, PBA, PHBookClub, phclassifieds, PHGamers, PHikingAndBackpacking, phlgbt, phr4r, Pilipinas, pinoyent, RedditPHCyclingClub, Tagalog, Tiangge, Gulong, dostscholars, AkoBaYungGago, Coronavirus\_PH, exIglesiaNiCristo, studentsph, medschoolph, lawstudentsph, DotA2, artph, FilipinoHistory, phclassifieds, Filipinology, CasualPH

\section{Data Sources}
\label{sec:data_sources}
The complete list of data sources are listed at Table \ref{tab:data_sources}.

\begin{table}[!ht]
\begin{tabularx}{\columnwidth}{|l|l|X|}
\hline
Domain & Source & URL \\ \hline
\multirow{7}{*}{News} & Abante & abante.com.ph \\ \cline{2-3} 
 & ABS-CBN & news.abs-cbn.com \\ \cline{2-3} 
 & Bandera & \begin{tabular}[c]{@{}l@{}}bandera.\\ inquirer.net\end{tabular} \\ \cline{2-3} 
 & GMA & gmanetwork.com \\ \cline{2-3} 
 & \begin{tabular}[c]{@{}l@{}}Manila\\ Bulletin\end{tabular} & mb.com.ph \\ \cline{2-3} 
 & \begin{tabular}[c]{@{}l@{}}Philippine\\ Star\end{tabular} & philstar.com \\ \cline{2-3} 
 & \begin{tabular}[c]{@{}l@{}}Radyo\\ Inquirer\end{tabular} & radyo.inquirer.net \\ \hline
\multirow{2}{*}{\begin{tabular}[c]{@{}l@{}}Social \\ Media\end{tabular}} & X/Twitter & twitter.com \\ \cline{2-3} 
 & YouTube & youtube.com \\ \hline
Encyclopedia & Wikipedia & \begin{tabular}[c]{@{}l@{}}dumps.wikimedia\\ .org/tlwiki/\\ latest\end{tabular} \\ \hline
\begin{tabular}[c]{@{}l@{}}Online \\ Forums\end{tabular} & Reddit & reddit.com \\ \hline
\multirow{3}{*}{Books} & \begin{tabular}[c]{@{}l@{}}Project \\ Gutenberg\end{tabular} & gutenberg.org \\ \cline{2-3} 
 & \begin{tabular}[c]{@{}l@{}}Google \\ Books\end{tabular} & books.google.com \\ \cline{2-3} 
 & Bible & wordproject.org \\ \hline
\end{tabularx}
\caption{List of sources per domain and its corresponding URLs.}
\label{tab:data_sources}
\end{table}

\section{Dynamic N-STEP vs Fixed N-STEP Clustering}
\label{sec:3step_experiment_section}

While Dynamic N-STEP Clustering allows for different number of clustering steps for each word, we want to know how it compares to its fixed counterpart. In this experiment, we compare Dynamic N-STEP and Fixed N-STEP clustering approach. We set \emph{N} to 3, which means there will be a fixed 3 clustering steps for each word, which we will refer to as 3-STEP Clustering from hereon. 

\begin{figure}[!ht]
    \begin{center}
    \includegraphics[width=\linewidth]{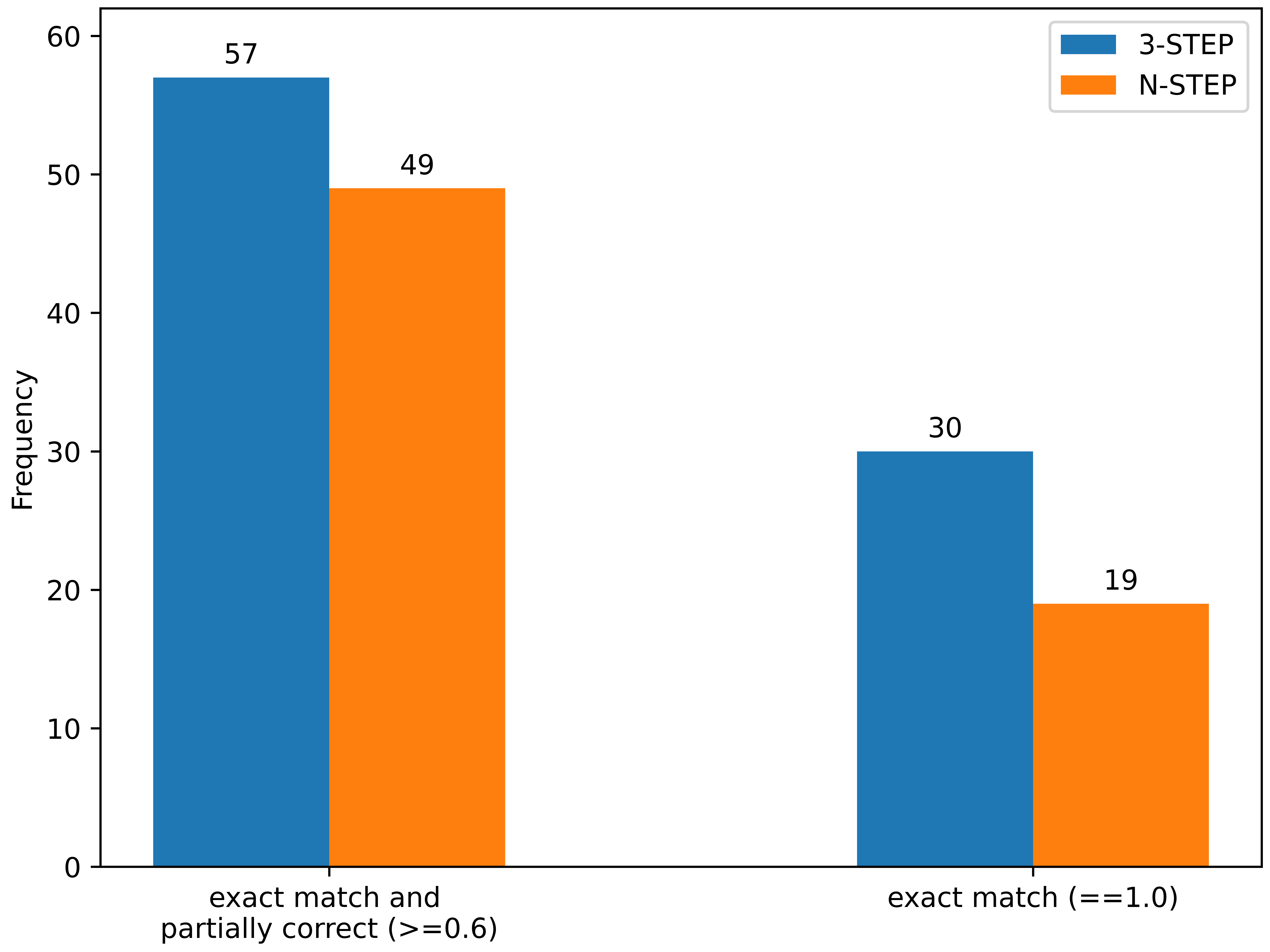} 
    \caption{The 3-STEP algorithm consistently outperforms Dynamic N-STEP algorithm in terms of the number of synsets perfectly matched and the number of synsets that are at least 60\% correct.}
    \label{fig:3step_vs_nstep_jaccard}
    \end{center}
\end{figure}

To compare both approaches, two independent wordnets were generated using the Dynamic N-STEP and 3-STEP clustering. The goal of this experiment is to see which approach can induce more synsets that matches with FilWordNet measured by Jaccard Index. The synset match counts were grouped into two groups, $>=0.6$ and $>=1.0$ (see Figure  \ref{fig:3step_vs_nstep_jaccard}). For both approaches, the same parameters was used as described in Section \ref{WSI_section}.

The results shows that the 3-STEP Clustering induced more correct synsets than Dynamic N-STEP Clustering. One possible explanation is that the larger sense count of 3-STEP Clustering (7,177 senses) compared to Dynamic N-STEP Clustering (5,452 senses) contributed to more correct and partially correct synsets, leading to higher coverage. The results suggests that, for our dataset, the 3-STEP clustering is much more effective than Dynamic N-STEP Clustering. However, this may not generalize to other datasets and we suggest using Dynamic N-STEP Clustering as a starting point for any datasets.

\begin{figure}[!ht]
    \begin{center}
    \includegraphics[width=\linewidth]{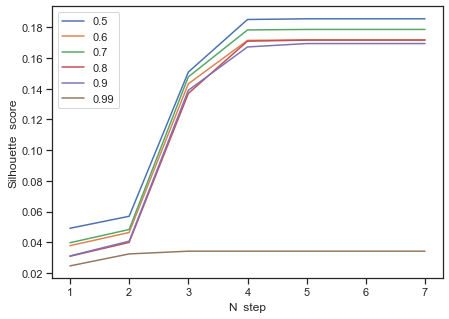} 
    \caption{The damping parameter of 0.5 shows consistently higher average silhouette scores throughout all clustering steps compared to other damping values. This shows that the ideal damping value is 0.5.}
    \label{fig:damping_experiment_silhouette}
    \end{center}
\end{figure}

\section{Setting the Damping Parameter}
\label{sec:damping_experiment_section}
The range of possible values for the damping parameter is 0.5 to 1, exclusive. Generally, the higher the damping, the lower the number of clusters. For example, setting the damping to 0.999 will result to just 1 cluster. To find the value for damping that maximizes the average silhouette score, different values were tested including 0.5, 0.6, 0.7, 0.8, 0.9, and 0.99. The clustering were tested on a fixed clustering steps for each word (7-STEP clustering) to observe the behavior of the silhouette score as the number of clustering steps increases. 

Figure \ref{fig:damping_experiment_silhouette} shows that damping = 0.5 consistently gives higher average silhouette scores for all clustering steps compared to other values. For damping = 0.99, it stopped improving the silhouette score at the second clustering step while all other values keeps improving until the fourth clustering step. With these results, we set the final damping parameter to 0.5.

\section{Setting the Distance Threshold}
\label{sec:distance_thres_experiment_section}
\begin{figure}[!ht]
    \begin{center}
    \includegraphics[width=\linewidth]{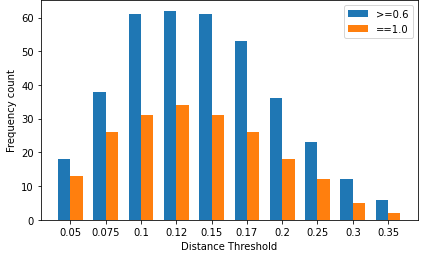} 
    \caption{The distribution of Jaccard Index per distance threshold shows that distance threshold of 0.12 obtains the most partial and exact matches.}
    \label{fig:jaccard_distribution}
    \end{center}
\end{figure}
One of the most important parameter of the Agglomerative Clustering algorithm is the distance threshold which decides whether to cluster or not. For synset induction, the ideal distance threshold is the one that maximizes the Jaccard Index of the induced synsets and FilWordNet synsets. To find that ideal threshold, different threshold values were tested (See Figure \ref{fig:jaccard_distribution}).

All of the synset entries that obtained a Jaccard Index of 0.6 or higher were counted. A Jaccard Index of 0.6 has been considered to give credit to partial correctness. Aside from this, Jaccard Index of 1 were also counted for comparison. The distance threshold of 0.12 achieved the highest frequency in terms of exact and partial matches of FilWordNet synsets for this particular dataset (see Figure \ref{fig:jaccard_distribution}). With these results, we set final cosine distance threshold parameter to 0.12.
\end{document}